\documentclass[conference]{IEEEtran}

\usepackage{amsmath}
\usepackage{amssymb} 

\usepackage[labelformat=simple]{subcaption}
\DeclareCaptionLabelSeparator{periodspace}{.\quad}
\usepackage{tikz}
\usetikzlibrary{arrows,matrix,positioning,patterns}
\usetikzlibrary{calc}
\usepackage{pgfplots} 
\usetikzlibrary{plotmarks}

\usepackage[utf8]{inputenc}
\usepackage{pgfplots}
\usepgfplotslibrary{groupplots}

\usepackage{url}
\usepackage{multirow}
\usepackage{colortbl}

\usepackage{algorithm} 
\usepackage{algpseudocode}

\DeclareSymbolFont{bbold}{U}{bbold}{m}{n}
\DeclareSymbolFontAlphabet{\mathbbold}{bbold}

\usepackage[labelformat=simple]{subcaption}
\DeclareCaptionLabelSeparator{periodspace}{.\quad}
\usepackage{listings}
\usepackage{xcolor}
\usepackage{cite}

\IEEEoverridecommandlockouts                              

\title{\LARGE \bf
From 2D to 3D terrain-following area coverage path planning}
\author{Mogens Plessen*
\thanks{*{\tt\small pmogens@proton.me}, Findklein GmbH, Switzerland}
}
\begin{document}

\maketitle
\thispagestyle{empty}
\pagestyle{empty}

%
%
%
%


\begin{abstract}
An algorithm for 3D terrain-following area coverage path planning is presented. Multiple adjacent paths are generated that are (i) locally apart from each other by a distance equal to the working width of a machinery, while (ii) simultaneously floating at a projection distance equal to a specific working height above the terrain. The complexities of the algorithm in comparison to its 2D equivalent are highlighted. These include uniformly spaced elevation data generation using an Inverse Distance Weighting-approach and a local search. Area coverage path planning results for real-world 3D data within an agricultural context are presented to validate the algorithm.
\end{abstract}



\section{Introduction\label{sec_intro}}

Agricultural real-world terrain is rarely completely flat. Therefore, 3D terrain data have to ideally be accounted for to generate precision paths for autonomous robotic area coverage path planning. For robots that operate within a field or work area with a specific working width and height, e.g. for spraying applications where nozzles are attached along a boombar that is carried by a vehicle at a specific working height  above the terrain (boom height), area coverage paths are sought that do neither produce area coverage gaps nor overlaps.

3D area coverage path planning is titularly not a new topic. However, a literature review revealed that there existed a research gap for a method that generates multiple adjacent paths that are (i) locally apart from each other by a distance equal to the working width of a machinery (e.g., the boombar width), while (ii) at the same time floating at each sampling point at a projection distance equal to a specific working height (e.g., the boom height) above the terrain. The presentation of such an algorithm is the contribution of this paper.

To the best of the author's knowledge this is for ground-based vehicles the first paper that presents an area coverage planning method over 3D terrain that simultaneously accounts for a specific working width and \emph{also} working height. 

For interest and extension, the standard commercial approach for area coverage path planning for uncrewed aerial vehicles (UAV)  in agriculture is to plan in 2D, before afterwards following the terrain via feedback control height reference tracking. This approach does also not simultaneously account for working width and height during the area coverage path planning step. 

\begin{figure}[htbp]
\centering
\begin{subfigure}[b]{0.5\textwidth}
\centering
\resizebox{\linewidth}{!}{
\begin{tikzpicture}
\draw[thick] (-3,1) .. controls (-1,1.2) and (1,1) .. (3,0) .. controls (5,-1) and (7,-1.2) .. (7,-1.2);
\fill[brown] (-1.2,1.05) circle (2pt);
\fill[brown] (3.6,-0.28) circle (2pt);
\draw [brown, dotted, thick] (-1.2,1.05) -- (3.6,-0.28);\draw[orange,thick,<->, >=latex'] (-1.3,0.65) -- (3.5,-0.68) node[midway, below] {$w$};
\end{tikzpicture}}
\caption{Visulization of adjacent paths resulting from the cylindical approach from \cite{hameed2016side}. Adjacent paths are spaced by a distance $w$.\vspace*{0.3cm}}
\label{fig_cylindr_subf1}
\end{subfigure}
%
%
\begin{subfigure}[b]{0.5\textwidth}
\centering
\resizebox{\linewidth}{!}{
\begin{tikzpicture}
%
%
%
%
%
\draw[thick] (-3,1) .. controls (-1,1.2) and (1,1) .. (3,0) .. controls (5,-1) and (7,-1.2) .. (7,-1.2);
\fill[brown] (-1.2,1.05) circle (2pt);
\fill[brown] (3.6,-0.28) circle (2pt);
\draw [brown, dotted, thick] (-1.2,1.05) -- (3.6,-0.28);\draw[orange,thick,<->, >=latex'] (-1.3,0.65) -- (3.5,-0.68) node[midway, below] {$w$};
\draw [blue, dashed, thick] (-1.2,1.05) -- (-1.25,2.05);
\draw [blue, dashed, thick] (3.6,-0.28) -- (4.01,0.69);
\fill[blue] (-1.26,2.08) circle (2pt);
\fill[blue] (4.0,0.66) circle (2pt);
\draw[blue,thick,<->, >=latex'] (-1.26,2.08) -- (4.0,0.66) node[midway, above] {$>w$};
\node[color=red] (a) at (-1.94, 1.59) {$h$};
\node[color=red] (a) at (4.1, 0.19) {$h$};
%
\draw[->,>=latex'] ($ (-0.6,1.05) + 0.3*({cos(180)},{sin(180)})$) arc (5:95:0.3);
\node[color=black,thick] (a) at (-1.08, 1.16) {$\cdot$};
%
\draw[->,>=latex'] ($ (3.9,-0.4)$) arc (-30:70:0.3);
\node[color=black,thick] (a) at (3.78,-0.21) {$\cdot$};
\draw[black,thick] (-1.55,1.065) -- (-1.43,1.06) -- (-1.45,1.65) -- (-1.57,1.64) -- (-1.55,1.05);
\draw[black,thick] (-1.5,1.4) -- (-1.52,1.85);
\draw[black,thick] (-1.75,1.85) -- (-0.75,1.85) -- (-0.75,2.35) -- (-1.75,2.35) -- (-1.75,1.85);
\draw[black,thick] (-0.83,1.05) -- (-0.95,1.06) -- (-0.97,1.62) -- (-0.85,1.63) -- (-0.83,1.05);
\draw[black,thick] (-0.9,1.38) -- (-0.92,1.85);
\end{tikzpicture}}        
\caption{The method implicitly assumes a working \emph{height} of zero. This simplification is in many agricultural applications not merited. E.g., the center-of-gravity (CoG) of some self-propelled sprayers may be more than 2m above ground.}
\label{fig_cylindr_subf2}
\end{subfigure}
\caption{Illustration of an important disadvantage of the method from \cite{hameed2016side}. A zero vehicle height is implicitly assumed. While adjacent paths are spaced by nominal working width $w$ along the ground surface, this is not the case anymore as soon as a vehicle height larger than zero, $h>0$, is assumed.}
\label{fig_cylindr}
\end{figure}
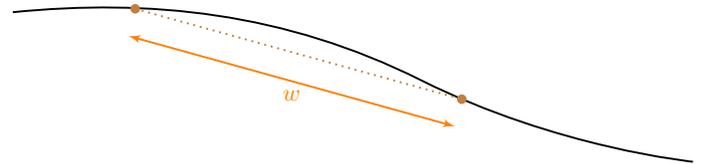
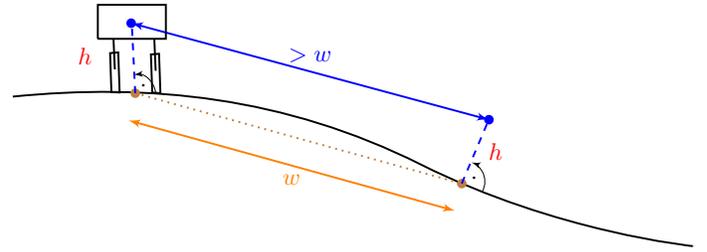

\begin{figure}
\centering
\begin{tikzpicture}
\node (image) at (0,0) {\includegraphics[width=6.5cm]{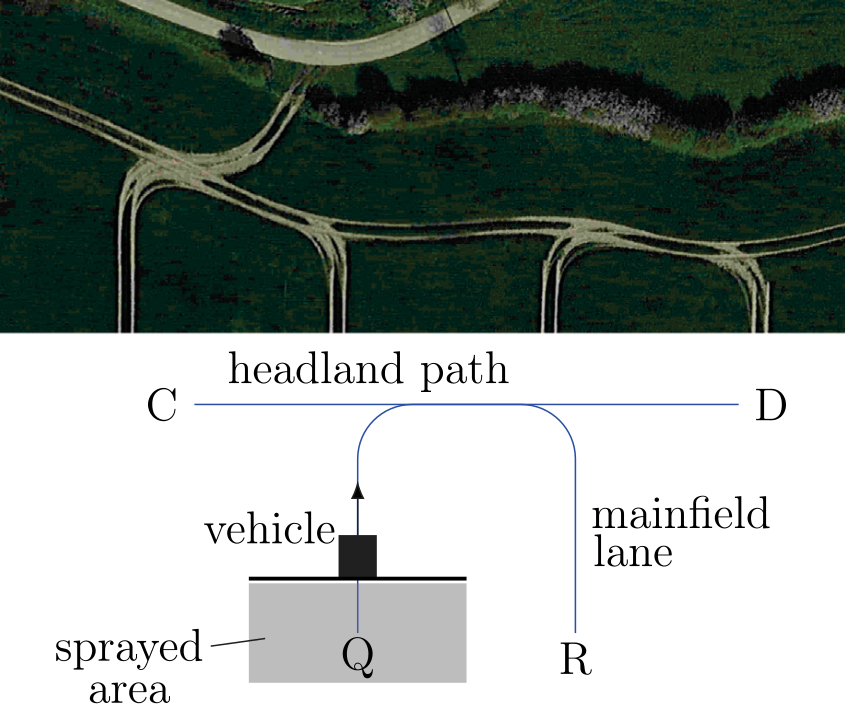}};
\node[color=black,font=\fontsize{9}{11}\selectfont] (a) at (-2.8, -1.4) {boombar};
\draw[black,line width=0.1pt] (-2.15,-1.5) -- (-1.1,-1.72);
\end{tikzpicture}

\caption{Bird's eye view: visualization of the terms \emph{headland path}, \emph{mainfield lane} and a vehicle covering an area with a working width to spray an area (e.g., pesticide application). \emph{Top}: Real-world visualization. \emph{Bottom}: Abstract visualization where a vehicle might travel from location Q towards R or D.}
\label{fig_problFormulation}
\end{figure}

\begin{figure}
\centering
\resizebox{\linewidth}{!}{
\begin{tikzpicture}
\draw[thick] (-3,1) .. controls (-1,1.2) and (1,1) .. (3,0) .. controls (5,-1) and (7,-1.2) .. (7,-1.2);
%
\draw [blue, dashed, thick] (-1.2,1.05) -- (-1.25,2.05);
\draw [blue, dashed, thick] (3.6,-0.28) -- (4.01,0.69);
\fill[blue] (-1.26,2.08) circle (2pt);
\fill[blue] (4.0,0.66) circle (2pt);
\draw[blue,thick,<->, >=latex'] (-1.26,2.08) -- (4.0,0.66) node[midway, above] {$w$};
\node[color=red] (a) at (-1.54, 1.59) {$h$};
\node[color=red] (a) at (4.1, 0.19) {$h$};
%
\draw[->,>=latex'] ($ (-0.6,1.05) + 0.3*({cos(180)},{sin(180)})$) arc (5:95:0.3);
\node[color=black,thick] (a) at (-1.08, 1.16) {$\cdot$};
%
\draw[->,>=latex'] ($ (3.9,-0.4)$) arc (-30:70:0.3);
\node[color=black,thick] (a) at (3.78,-0.21) {$\cdot$};
\draw[blue,thick,<-, >=latex'] (-1.26,2.08) -- (-3.1,1.9);
\node[color=blue] (a) at (-2.43, 2.18) {$w$};
\draw[blue,thick,<-, >=latex'] (4.0,0.66) -- (7.1,-0.36);
\node[color=blue] (a) at (6., 0.22) {$w$};
\node[rotate=-19,color=black] (a) at (1.8, -0.02) {$terrain$};
\draw[black,thick] (-2.5,0.5) -- (-2,0.75);
\draw[black,thick] (-1.5,0.5) -- (-1,0.75);
\draw[black,thick] (-0.5,0.4) -- (-0,0.65);
\draw[black,thick] (0.3,0.2) -- (0.8,0.45);
\draw[black,thick] (2.7,-0.6) -- (3.2,-0.35);
\draw[black,thick] (3.7,-1.0) -- (4.2,-0.75);
\draw[black,thick] (4.7,-1.2) -- (5.2,-0.95);
\end{tikzpicture}
}
\caption{Goal: area coverage paths floating at a projection distance $h$ above the terrain while adjacent mainfield lanes are spaced by distance $w$.}
\label{fig_sketchobj}
\end{figure}
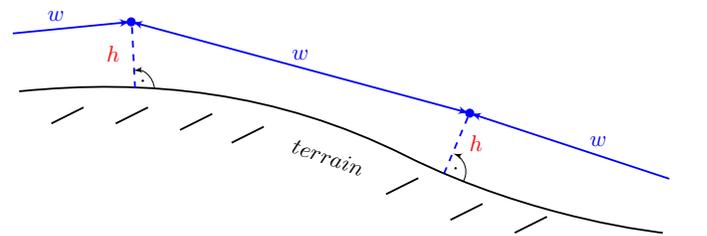

The remaining paper focuses exclusively on the case for ground-based vehicles. A literature review is provided. The standard approach to extend 2D area coverage path planning methods to the 3D case is to plan in 2D before projecting the path onto the terrain via a Digital Elevation Model (DEM) that maps, e.g., latitude and longitude coordinates to elevation coordinates. However, while on the latitude-longitude level two adjacent paths may appear parallel to each other for a specific working width, they are not for topographically varying terrain. A plethora of papers exist that maintain this general projection approach but search for an optimal 'driving direction' that accounts for 3D terrain data, and often optimize a multi-objective function involving energy consumption \cite{hameed2014intelligent, dogru2015energy, shen2020simulation,  bostelmann2023multi, vahdanjoo2024three, jiang2024three, lin2025path}. The search for the optimal driving direction is often done stochastically via genetic programming \cite{qiu2024terrain}. Multiple papers also propose the partition the field area into several subfields to exploit varying terrain characteristics \cite{jin2011coverage, vreznik2021towards, pour20253d}. Specific path shaping mechanisms were considered. 
A Fermat spiral path planning method to exploit different height contour lines is used in \cite{wu2019energy}. Constraints on trajectory curvature are taken into account via second-order cone programming \cite{tormagov2021coverage}. Beyond agricultural fields the topic of 3D path planning is also relevant for lawn mowers \cite{zhou2025coverage} and vineyards \cite{contente2015path,santos2018path}.

Closest to the objective of the present paper is the work of \cite{hameed2016side}, where a `\emph{3D side-to-side coverage algorithm is proposed in which spaces between adjacent swaths are kept equal taking into account the topographical nature of the field}'. To create adjacent swaths it is proposed to (i) create at each sampling point and for a specific heading a cylinder with radius equal to the working width, before (ii) intersecting this cylinder with the 3D topographical surface to obtain the corresponding coordinate of an adjacent swath. This approach, however, has an important disadvantage. It implicitly assumed a zero vehicle height. See Fig. \ref{fig_cylindr} for visualization.

The remaining paper is organized as follows: problem formulation, problem solution, numerical examples, discussion and the conclusion are described in Sect. \ref{sec_intro}-\ref{sec_conclusion}.

\section{Problem Formulation\label{sec_probformul}}

A robot with a working width (boombar width) of $w$ is assumed to be operating in an agricultural field or, more generally, a specific work area. The boombar shall be maintained at a nominal boom height $h$ above the terrain. Efficient path plans are desired for full area coverage. 

Planning in 2D, typically based on latitude and longitude data, implicitly assumes a flat terrain. When tracking a 2D area coverage path plan in real-world non-planar terrain, spray \emph{gaps} result. One solution to avoid spray gaps while maintaining a 2D planning technique is to determine the maximum spray gap distance from data, reducing the effective working width by this distance, before recalculating a new area coverage path plan for this effective (instead of the original) working width. While the resulting path would not generate any spray gaps, however, instead now spray \emph{overlaps} would result.

Thus, to minimize both spray gaps as well as overlaps it has to be accounted for 3D terrain data directly during area coverage path planning. To provide such a method is the problem addressed in this paper. See Fig. \ref{fig_problFormulation} and \ref{fig_sketchobj} for illustration.

Note that \emph{partial} field coverage (e.g., as a result of accounting for varying spray tank level dynamics during field operations) and the like are not of concern for this paper, where the focus is on providing a mathematical algorithm to compute area coverage paths accounting for 3D terrain data.

\section{Problem Solution\label{sec_probsoln}}

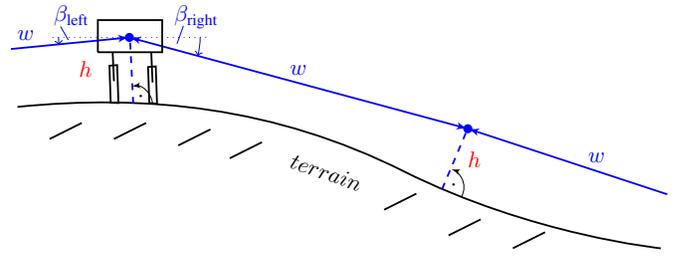
\begin{figure}
\centering
\resizebox{\linewidth}{!}{
\begin{tikzpicture}
\draw[thick] (-3,1) .. controls (-1,1.2) and (1,1) .. (3,0) .. controls (5,-1) and (7,-1.2) .. (7,-1.2);
%
\draw [blue, dashed, thick] (-1.2,1.05) -- (-1.25,2.05);
\draw [blue, dashed, thick] (3.6,-0.28) -- (4.01,0.69);
\fill[blue] (-1.26,2.08) circle (2pt);
\fill[blue] (4.0,0.66) circle (2pt);
\draw[blue,thick,<->, >=latex'] (-1.26,2.08) -- (4.0,0.66) node[midway, above] {$w$};
\node[color=red] (a) at (-1.94, 1.59) {$h$};
\node[color=red] (a) at (4.1, 0.19) {$h$};
%
\draw[->,>=latex'] ($ (-0.6,1.05) + 0.3*({cos(180)},{sin(180)})$) arc (5:95:0.3);
\node[color=black,thick] (a) at (-1.08, 1.16) {$\cdot$};
%
\draw[->,>=latex'] ($ (3.9,-0.4)$) arc (-30:70:0.3);
\node[color=black,thick] (a) at (3.78,-0.21) {$\cdot$};
\draw[blue,thick,<-, >=latex'] (-1.26,2.08) -- (-3.1,1.9);
\node[color=blue] (a) at (-2.87, 2.12) {$w$};
\draw[blue,thick,<-, >=latex'] (4.0,0.66) -- (7.1,-0.36);
\node[color=blue] (a) at (6., 0.22) {$w$};
\node[rotate=-19,color=black] (a) at (1.8, -0.02) {$terrain$};
\draw[black,thick] (-2.5,0.5) -- (-2,0.75);
\draw[black,thick] (-1.5,0.5) -- (-1,0.75);
\draw[black,thick] (-0.5,0.4) -- (-0,0.65);
\draw[black,thick] (0.3,0.2) -- (0.8,0.45);
\draw[black,thick] (2.7,-0.6) -- (3.2,-0.35);
\draw[black,thick] (3.7,-1.0) -- (4.2,-0.75);
\draw[black,thick] (4.7,-1.2) -- (5.2,-0.95);
\draw[black,thick] (-1.55,1.065) -- (-1.43,1.06) -- (-1.45,1.65) -- (-1.57,1.64) -- (-1.55,1.05);
\draw[black,thick] (-1.5,1.4) -- (-1.52,1.85);
\draw[black,thick] (-1.75,1.85) -- (-0.75,1.85) -- (-0.75,2.35) -- (-1.75,2.35) -- (-1.75,1.85);
\draw[black,thick] (-0.83,1.05) -- (-0.95,1.06) -- (-0.97,1.62) -- (-0.85,1.63) -- (-0.83,1.05);
\draw[black,thick] (-0.9,1.38) -- (-0.92,1.85);
\draw[blue,dotted] (-1.26,2.08) -- (-2.51,2.08);
\draw[->,blue] ($ (-1.26,2.08) + 1.1*({cos(180)},{sin(180)})$) arc (180:186:1.1);
\node[color=blue] (a) at (-2.16,2.42) {$\beta_\text{left}$};
\draw [blue] (-2.33,2.25) -- (-2.21,2.04);
\draw[blue,dotted] (-1.26,2.08) -- (-0.01,2.08);
\draw[->,blue] ($ (-1.26,2.08) + 1.1*({cos(0)},{sin(0)})$) arc (0:-15:1.1);
\node[color=blue] (a) at (-0.22,2.4) {$\beta_\text{right}$};
\draw [blue] (-0.42,2.26) -- (-0.52,1.98);
\end{tikzpicture}
}
\caption{(i) Vehicle visualization, and (ii) illustration of the assumption that boombar halves can be controlled independently up to a specific angle. The control of two boombar halves over two adjacent paths enables full working width $w$.}
\label{fig_vehicle}
\end{figure}

Let 3D position coordinates be denoted in the classic three-dimensional Cartesian coordinate system by $(x,y,z)\in\mathbb{R}^3$, where the $z$-coordinate denotes elevation. Let $\beta_j\in(-\pi/2,\pi/2)$ for $j\in\{\text{left},\text{right}\}$ denote the angles of the boombar with respect to sea-level ($x-y$ plane) on the left and right-hand side of the center of the boombar, respectively. It is assumed that these angles can be controlled independently. See Fig. \ref{fig_vehicle} for illustration. Let $\Delta h(b)>0,\forall b\in[-w/2,w/2]$ denote the boom height with respect to the terrain as a function of position $b$ along the boombar. 

The core necessity for efficient area coverage path planning is to generate sub-paths, here called \emph{mainfield lanes} (see Fig. \ref{fig_problFormulation}) that are locally minimizing (and ideally eliminating) both spray gaps as well as overlaps. In agriculture, where the work area is typically divided into a headland area used for turning and a mainfield area, this concept is predominantly important for the path planning of mainfield lanes. However, it may also be relevant for headland area paths, e.g., if multiple adjacent headland paths are planned.  

To summarize, the main difficulty for 3D terrain-following area coverage path planning is to efficiently generate locally adjacent paths that maintain a distance of the working width $w$ to each other, while accounting for a working height $h$ and the 3D terrain.

 Note that once a 3D lane grid is created that connects all mainfield lanes and headland paths, a graph optimization can be solved applying the exact same techniques as for 2D area coverage path planning (e.g. \cite{plessen2019optimal}).

\begin{figure}
\centering
\resizebox{\linewidth}{!}{
\begin{tikzpicture}
\draw[thick] (3,1) .. controls (1,1.2) and (-1,1) .. (-3,0) .. controls (-5,-1) and (-7,-1.2) .. (-7,-1.2);
\fill[brown] (1.2,1.05) circle (2pt);
\fill[brown] (-3.6,-0.28) circle (2pt);
\draw [blue, dashed, thick] (1.2,1.05) -- (1.11,2.06);
\draw [blue, dashed, thick] (-3.6,-0.28) -- (-4.01,0.69);
\fill[blue] (1.11,2.06) circle (2pt);
\fill[blue] (-4.0,0.66) circle (2pt);
\draw[blue,thick,<->, >=latex'] (1.11,2.06) -- (-4.0,0.66) node[midway, above] {$w$};
\node[color=blue] (a) at (0.91, 1.46) {$h$};
\node[color=blue] (a) at (-3.55, 0.33) {$h$};
%
\draw [black, dotted] (0.2,1.05) -- (2.2,1.05);
\draw [black, dotted] (1.2,0.05) -- (1.2,2.55);
\draw[->,red] ($ (1.2,1.05) + 0.74*({cos(90)},{sin(90)})$) arc (90:95:0.74);
\node[color=red] (a) at (1.53, 1.72) {$\alpha$};
\draw [red] (1.37,1.7) -- (1.17,1.62);
\draw[->,red] ($ (1.2,1.05) + 0.94*({cos(180)},{sin(180)})$) arc (180:185:0.94);
\draw [red] (0.58,0.87) -- (0.35,1.03);
\node[color=red] (a) at (0.7, 0.8) {$\alpha$};
%
\draw [black, dotted] (-5.3,2.06) -- (2.11,2.06);
\draw[->,>=latex'] ($ (1.11,2.06) + 1.3*({cos(180)},{sin(180)})$) arc (180:195:1.3);
\node[color=black] (a) at (0.43,2.25) {$\beta$};
\draw [black] (0.27,2.2) -- (0.03,1.91);
\draw[->,>=latex',dotted] ($ (1.11,2.06) + 5.3*({cos(180)},{sin(180)})$) arc (180:195:5.3);
\draw[dotted] ($ (1.11,2.06) + 5.3*({cos(180)},{sin(180)})$) arc (180:175:5.3);
\node[color=black] (a) at (-2.81, 2.25) {$r=w$};
\draw[->,>=latex'] ($ (-3.6,-0.28) + 0.2*({cos(25)},{sin(25)})$) arc (25:115:0.2);
\node[color=black] (a) at (-3.56,-0.16) {$\cdot$};
\draw [blue,dotted] (-4.0,1.2) -- (-4.0,-0.66);
\draw[->,>=latex',blue] ($ (-4.0,0.66) + 0.56*({cos(-90)},{sin(-90)})$) arc (-90:-66:0.56);
\node[color=blue] (a) at (-4.3,0.35) {$\gamma$};
\draw [blue] (-4.2,0.3) -- (-3.94,0.25);
\draw [->,>=latex',thick,gray] (-6.7,-0.7) -- (-6.7,-0.1);
\node[color=black,gray] (a) at (-6.48,-0.47) {$z$};
\end{tikzpicture}}
\caption{Visualization of variables used in the Algorithm of Sect. \ref{subsec_algo}.}
\label{fig_gamma}
\end{figure}
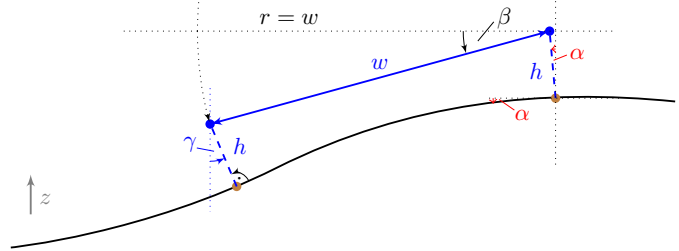

\subsection{Generating adjacent paths in 3D terrain \label{subsec_3dpath}}

In the following a method is described to generate adjacent mainfield lanes for area coverage path planning along 3D terrain. 

Assume a 3D reference path (e.g., a previously generated mainfield lane) is available, $\{(x_i^{(k)},y_i^{(k)},z_i^{(k)})\}_{i=0}^{N^{(k)}}$, that is described by an ordered sequence of $N^{(k)}>0$ coordinates and where the mainfield lane number shall be indexed by $k>0$. An adjacent mainfield lane path, $\{(x_i^{(k+1)},y_i^{(k+1)},z_i^{(k+1)})\}_{i=0}^{N^{(k+1)}}$, is sought. Concatenating multiple of such adjacent lanes shall produce area coverage, while minimizing spray gaps as well as overlaps along 3D terrain. 

The 3D terrain is assumed to be available in the form of a look-up table in combination with an interpolation routine that maps $(x,y)$-coordinates to elevation, $z=f(x,y),\forall x\in\mathcal{X},y\in\mathcal{Y}$, where $\mathcal{X}$ and $\mathcal{Y}$ denote the $x$- and $y$-coordinate ranges relevant for a given field area, respectively. The construction of an efficient look-up table is discussed further below in Sect. \ref{subsec_ztable}.

\subsection{Algorithm\label{subsec_algo}}

For each coordinate along mainfield lane $k$, the following calculations are computed. First,
\begin{subequations}
\begin{align}
\psi_i^{(k)} &= \arctan ( \frac{y_{i+1}^{(k)}-y_{i}^{(k)}}{x_{i+1}^{(k)}-x_{i}^{(k)}} ),\\
\begin{bmatrix} x_{i,(m)}^{(k)}\\ y_{i,(m)}^{(k)} \\ z_{i,(m)}^{(k)} \end{bmatrix} &= \frac{1}{2}\begin{bmatrix} x_{i}^{(k)}+x_{i+1}^{(k)}\\ y_{i}^{(k)}+y_{i+1}^{(k)}\\ z_{i}^{(k)}+z_{i+1}^{(k)} \end{bmatrix},\label{eq_xyzm}
\end{align}
\end{subequations}

Second, if the point \eqref{eq_xyzm} is elevated at a projection distance of approximately the nominal boom height $h$ above the terrain, set: 
\begin{equation}
\begin{bmatrix} x_{i,(h)}^{(k)}\\ y_{i,(h)}^{(k)} \\ z_{i,(h)}^{(k)} \end{bmatrix} = \begin{bmatrix} x_{i,(m)}^{(k)}\\ y_{i,(m)}^{(k)} \\ z_{i,(m)}^{(k)} \end{bmatrix},
\end{equation}
else, calculate: 
\begin{subequations}
\begin{align}
\begin{bmatrix} x_{i,(a)}^{(k)}\\ y_{i,(a)}^{(k)} \end{bmatrix} &= \begin{bmatrix} x_{i,(m)}^{(k)}\\ y_{i,(m)}^{(k)} \end{bmatrix} + a \begin{bmatrix} \cos(\psi_i^{(k)}+\frac{\pi}{2})\\ \sin(\psi_i^{(k)}+\frac{\pi}{2}) \end{bmatrix},\label{eq_1stset_1st}\\
z_{i,(a)}^{(k)} &= f(x_{i,(a)}^{(k)},y_{i,(a)}^{(k)}),\label{eq_f_1}\\
\alpha_{i,(a)}^{(k)} &= \arctan ( \frac{z_{i,(m)}^{(k)} - z_{i,(a)}^{(k)} }{a} ),\\
\begin{bmatrix} x_{i,(h)}^{(k)}\\ y_{i,(h)}^{(k)} \end{bmatrix} &= \begin{bmatrix} x_{i,(m)}^{(k)}\\ y_{i,(m)}^{(k)} \end{bmatrix} + h \sin(\alpha_{i,(a)}^{(k)}) \begin{bmatrix} \cos(\psi_i^{(k)}+\frac{\pi}{2})\\ \sin(\psi_i^{(k)}+\frac{\pi}{2}) \end{bmatrix},\\
z_{i,(h)}^{(k)} &= f(x_{i,(m)}^{(k)},y_{i,(m)}^{(k)}) + h\cos(\alpha_{i,(a)}^{(k)}),\label{eq_f_2}
\end{align}
\label{eq_1stset}
\end{subequations}
where $a>0$ denotes half of the tractor vehicle's axle width.

Third,
\begin{subequations}
\begin{align}
\begin{bmatrix} x_{i,(w)}^{(k+1)}\\ y_{i,(w)}^{(k+1)} \end{bmatrix} &= \begin{bmatrix} x_{i,(h)}^{(k)}\\ y_{i,(h)}^{(k)} \end{bmatrix} + w \begin{bmatrix} \cos(\psi_i^{(k)}+\frac{\pi}{2})\\ \sin(\psi_i^{(k)}+\frac{\pi}{2}) \end{bmatrix},\label{eq_2ndset_1st}\\
z_{i,(w)}^{(k+1)} &= f(x_{i,(w)}^{(k+1)},y_{i,(w)}^{(k+1)}),\label{eq_f_3}\\
\beta_{i,(w)}^{(k+1)} &= \arctan (  \frac{z_{i,(h)}^{(k)} - z_{i,(w)}^{(k+1)} - h }{w} ),\label{eq_2ndset_beta}\\
\hat{w}_{i,(w)}^{(k+1)} &= w \cos(\beta_{i,(w)}^{(k+1)}),\label{eq_what}\\
\begin{bmatrix} x_{i,(w)}^{(k+1)}\\ y_{i,(w)}^{(k+1)} \end{bmatrix} &= \begin{bmatrix} x_{i,(h)}^{(k)}\\ y_{i,(h)}^{(k)} \end{bmatrix} + \hat{w}_{i,(w)}^{(k+1)} \begin{bmatrix} \cos(\psi_i^{(k)}+\frac{\pi}{2})\\ \sin(\psi_i^{(k)} +\frac{\pi}{2}) \end{bmatrix},\\
z_{i,(w)}^{(k+1)} &= z_{i,(h)}^{(k)} - w \sin(\beta_{i,(w)}^{(k+1)}) ,\label{eq_2ndset_zinterm}\\
d_{i,(w)}^{\text{vert},(k+1)} &= z_{i,(w)}^{(k+1)} - f(x_{i,(w)}^{(k+1)}, y_{i,(w)}^{(k+1)}) ,\label{eq_2ndset_dvert}\\
h_{i,(w)}^{\text{proj},(k+1)} &= | d_{i,(w)}^{\text{vert},(k+1)} \cos(\beta_{i,(w)}^{(k+1)}) |,\label{eq_2ndset_hproj}
\end{align}
\label{eq_2ndset}
\end{subequations}
where $h_{i,(w)}^{\text{proj},(k+1)}$ approximates the projection distance to the terrain. This approximation is done (i) for computational efficiency (to avoid an additional local search), and (ii) is reasonable for typically small roll angles $\beta_{i,(w)}^{(k+1)}$. A more accurate alternative would require an additional local search to determine angle $\gamma$ according to Fig. \ref{fig_gamma} to calculate the correct projection distance since the terrain is in general arbitrarily locally varying,\\ $ h_{i,(w)}^{\text{proj},(k+1)}= \min_{\substack{
    x\in\mathcal{X},\\
    y\in\mathcal{Y},\\
    z=f(x,y)
}}  \{|| \begin{bmatrix} x_{i,(w)}^{(k+1)}\\ y_{i,(w)}^{(k+1)} \\ z_{i,(w)}^{(k+1)} \end{bmatrix} - \begin{bmatrix} x\\ y \\ z \end{bmatrix} ||_2 \}$,\\ where the projection distance is calculated using the Euclidean norm.

Fourth, if $|h_{i,(w)}^{\text{proj},(k+1)}-h|<\epsilon$, where $\epsilon>0$ is a hyperparameter small tolerance level, return 
\begin{equation}
\begin{bmatrix} x_{i}^{(k+1)}\\ y_{i}^{(k+1)} \\ z_{i}^{(k+1)} \end{bmatrix} = \begin{bmatrix} x_{i,(w)}^{(k+1)}\\ y_{i,(w)}^{(k+1)} \\ z_{i,(w)}^{(k+1)} \end{bmatrix},\label{ex_xyzikp1}
\end{equation}
else, use a hyperparameter angle, $\Delta \beta>0$, to adapt
\begin{equation}
\beta_{i,(w)}^{(k+1)} = \begin{cases} \beta_{i,(w)}^{(k+1)} - \Delta \beta,~\text{if}~h_{i,(w)}^{\text{proj},(k+1)}<h,\\
\beta_{i,(w)}^{(k+1)} + \Delta \beta,~\text{else},
 \end{cases}\label{eq_beta_sear}
\end{equation}
and repeat Steps \eqref{eq_what}-\eqref{eq_2ndset_hproj} until, $|h_{i,(w)}^{\text{proj},(k+1)}-h|\leq\epsilon$, or the tolerance error does not further decrease, and \eqref{ex_xyzikp1} is returned.


\subsection{Details and extensions}

Multiple comments are made. First, hyperparameters used in the evaluation experiments are listed in Table \ref{tab_hyperparam}. They are intuitive and well interpretable as angle stepsize and distance tolerance.

\begin{table}
\centering
\caption{Hyperparameters used in numerical experiments, including angle stepsize and distance tolerance..  \label{tab_hyperparam}}
 \def\arraystretch{1.0}
 \begin{tabular}{|l|l|}
 \hline
Hyperp. & Value \\
\hline
$\epsilon$ & 0.1 (m) \\
$\Delta \beta$ & 1 (°) \\
\hline 
\end{tabular}
\vspace{-0.3cm}
\end{table}


Second, the stopping criterion for the iterations of Steps \eqref{eq_what}-\eqref{eq_2ndset_hproj} is specified  when the tolerance error does not further decrease. Let $h_{i,(w)}^{\text{proj},(k+1)}(j-1)$ and $h_{i,(w)}^{\text{proj},(k+1)}(j)$ denote the projection distances at two consecutive iteration indices $j-1$ and $j$, respectively. 

Then, if $\text{sign}( h_{i,(w)}^{\text{proj},(k+1)}(j)-h )\neq \text{sign}( h_{i,(w)}^{\text{proj},(k+1)}(j-1)-h )$, compute
\begin{subequations}
\begin{align}
\eta_j &= \frac{1}{|h_{i,(w)}^{\text{proj},(k+1)}(j)-h|},\label{eq_omega1}\\
\eta_{j-1} &= \frac{1}{|h_{i,(w)}^{\text{proj},(k+1)}(j-1)-h|},\label{eq_omega2}\\
\beta_{i,(w)}^{(k+1)} &= \frac{\eta_j \beta_{i,(w)}^{(k+1)}(j) + \eta_{j-1} \beta_{i,(w)}^{(k+1)}(j-1) }{\eta_j+\eta_{j-1}},\\
\hat{w}_{i,(w)}^{(k+1)} &= w \cos(\beta_{i,(w)}^{(k+1)}),\\
\begin{bmatrix} x_{i,(w)}^{(k+1)}\\ y_{i,(w)}^{(k+1)} \end{bmatrix} &= \begin{bmatrix} x_{i,(h)}^{(k)}\\ y_{i,(h)}^{(k)} \end{bmatrix} + \hat{w}_{i,(w)}^{(k+1)} \begin{bmatrix} \cos(\psi_i^{(k)}+\frac{\pi}{2})\\ \sin(\psi_i^{(k)} +\frac{\pi}{2}) \end{bmatrix},\label{eq_omega_xy}\\
z_{i,(w)}^{(k+1)} &= z_{i,(h)}^{(k)} - w \sin(\beta_{i,(w)}^{(k+1)}),\label{eq_omega_z}
\end{align}
\label{eq_omega}
\end{subequations}
before terminating and returning the results in \eqref{eq_omega_xy} and \eqref{eq_omega_z} as \eqref{ex_xyzikp1}.

Third, note that as a result of above computations adjacent paths result that are locally apart from each other by distance $w$, while at the same time floating at each sampling point at a projection distance of approximately $h$ above the terrain.

Fourth, note that the nominal boom height or projection distance $h$ is enforced (subject to tolerance $\epsilon$) only at mainfield lane centerlines. There is in general no guarantee that this distance is maintained along the entire width of the boombar. In fact, over uneven terrain it is impossible to ensure this along the entire width of the boombar for the general case of arbitrarily locally varying terrain. This is an important observation. Each left and right boombar half is modeled as a rigid body. We selected projection distance to be $h$ at the outest boombar tips (i.e., at the CoG mainfield lane paths), respectively. Other design selections, e.g. at the halves of each boombar half are also possible.

Finally, during the computation of above Algorithm the condition, $|\beta_{i,(w)}^{(k+1)}|<\beta_\text{boombar}^\text{max}$, can be used to exclude area coverage paths resulting in excessive roll angles or roll angles above a threshold $\beta_\text{boombar}^\text{max}>0$. If a path with a roll angle exceeding the threshold is detected, a new alternative initial reference path (e.g., an alternative subsection of the area contour) is selected before the Algorithm of Sect. \ref{subsec_algo}  is restarted.

%
%

\subsection{Look-up table preparation for elevation computation \label{subsec_ztable}}

3D terrain data is assumed to be available in the form of a look-up table. A fast evaluation of \eqref{eq_f_1},  \eqref{eq_f_2}, \eqref{eq_f_3} and \eqref{eq_2ndset_dvert} is crucial for overall computational efficiency of above Algorithm. One approach is to organize the look-up table based on a uniform interpolation grid. The steps are as follows. First, suppose sampled 3D terrain data is initially available arbitrarily non-uniformly spaced, $\{x_{j,(s)},y_{j,(s)},z_{j,(s)}\}_{j=0}^{N_{(s)}}$, with $N_{(s)}>0$ data points. Second, decide a hyperparameter interpolation grid spacing, $\Delta g>0$, used for both $x$- and $y$-axes. Third, prepare the look-up table by evaluating the elevation along the interpolation grid using an \emph{Inverse Distance Weighting} (IDW) approach \cite{shepard1968two}, 
\begin{equation}
f(x_m,y_n) = \begin{cases} z_q^{\star,(m,n)},~\text{if}~d_q^{\star,(m,n)}=0,\\
\frac{ \sum_{q=0}^{Q_\text{IDW}} \eta_q^{\star,(m,n)} z_q^{\star,(m,n)} }{ \sum_{q=0}^{Q_\text{IDW}} \eta_q^{\star,(m,n)} },~\text{else},\end{cases}
\end{equation}
where $\eta_q^{\star,(m,n)}=\frac{1}{d_q^{\star,(m,n)}}$ for $d_q^{\star,(m,n)}=\sqrt{(x_m-x_q^{\star})^2 + (y_n-y_q^{\star})^2}$, and indices $q\in\{0,\dots,Q_\text{IDW}\}$ indicate the $Q_\text{IDW}+1$ in Euclidean distance measured closest points to the interpolation grid coordinate $(x_m,y_n)$. In experiments hyperparameters are selected as $\Delta g=1$m and $Q_\text{IDW}=3$.

Once 3D terrain data is available over an uniformly spaced interpolation grid, classic \emph{Bilinear Interpolation} (BI) can be used to interpolate the elevation at an arbitrary $(x,y)$-coordinate. Suppose $(x,y)$ is within a cell of the uniform interpolation grid, such that $x_l\leq x\leq x_u$ and $y_l\leq y\leq y_u$ and $f(x_j,y_j),\forall j\in\{l,u\}$ is available, then $f(x,y)$ can be evaluated as follows:
\begin{subequations}
\begin{align}
f(x,y_l) &= \frac{x_u-x}{x_u-x_l}f(x_l,y_l)+\frac{x-x_l}{x_u-x_l}f(x_u,y_l),\\
f(x,y_u) &= \frac{x_u-x}{x_u-x_l}f(x_l,y_u)+\frac{x-x_l}{x_u-x_l}f(x_u,y_u),\\
f(x,y) &= \frac{y_u-y}{y_u-y_l}f(x,y_l)+\frac{y-y_l}{y_u-y_l}f(x,y_u).
\end{align}
\end{subequations}

Bilinear interpolation cannot be used directly on raw data since it is in general non-uniformly spaced by above assumption. This necessitated above IDW-step first.

\subsection{Path tangential interpolation spacing \label{subsec_dspacing}}

All of above discussion focused on details for the creation of adjacent paths while accounting for 3D terrain data, the working width $w$ and height $h$. Above discussion did not yet address path tangential interpolation spacings, \begin{small}$d_i^{(k)}=\sqrt{ (x_{i+1}^{(k)} - x_i^{(k)})^2 + (y_{i+1}^{(k)} - y_i^{(k)})^2 + (z_{i+1}^{(k)} - z_i^{(k)})^2}$\end{small},  for $i=0,\dots,N^{(k)}$ coordinates along each of the adjacent paths $k=1,\dots,K$. 

When generating a cascaded sequence of multiple adjacent paths, the interpolation distance along the first reference path is highly influential on the result of all the adjacent paths. This is since any potential projection errors are accumulated over multiple projections to generate the multiple adjacent paths. 

Selecting a specific interpolation spacing along the reference path is a heuristic choice. The method suggested in this paper, and used for the numerical experiments in Sect. \ref{sec_expts}, is (i) based on only 2 parameters, namely working width $w$ and an assumed maximal permissible absolute heading change $\Delta \psi_\text{max}>0$ along the reference path, and (ii) is derived based on geometric argumentation to avoid generation of jaggedness in adjacent paths. The guiding inequality is
\begin{equation*}
d \text{sin}(\Delta \psi_\text{max}) + w \text{cos}(\Delta \psi_\text{max}) \geq w,
\end{equation*}
from which the minimum path tangential interpolation spacing is calculated as, $d \geq w(1-\text{cos}(\Delta \psi_\text{max}))/\text{sin}(\Delta \psi_\text{max})$. 

Assuming a maximal permissible $\Delta \psi_\text{max}=30^\circ$ and working width $w=36$m, this results in $d_\text{min}\approx 10$m. In below numerical experiments, this interpolation spacing is enforced along the initial reference path. 

It is underlined that above method is \emph{one} possible heuristic. More complex ones, e.g. that also account for height, maximum absolute roll angle assumptions or are fully data-dependent based on (exhaustive) terrain searches, are possible. The benefit of above method with only 2 parameters is its simplicity.

\section{Numerical Experiments\label{sec_expts}}

\begin{figure}
\centering
\includegraphics[width=8.7cm]{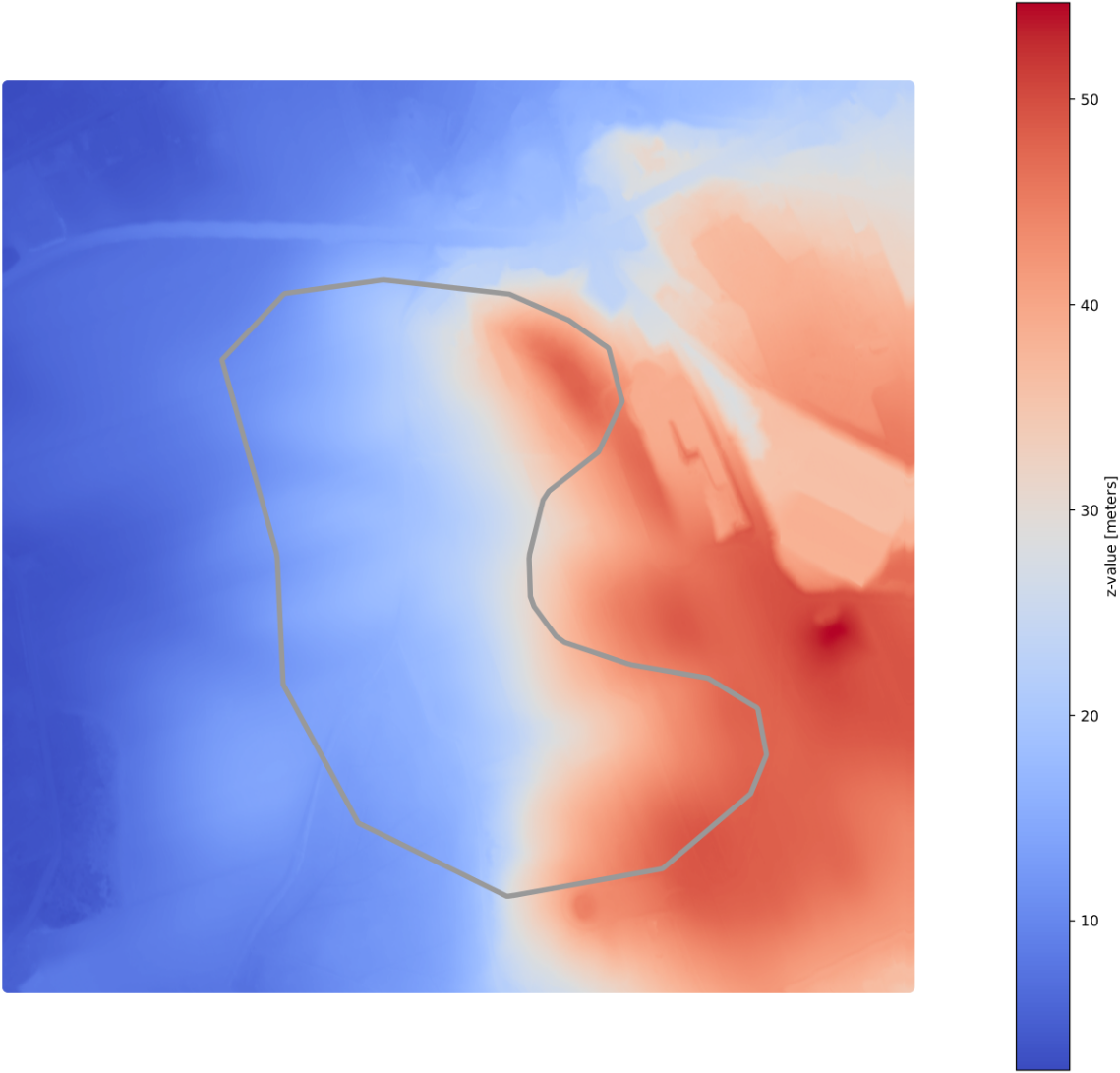}
\caption{Illustration of data used for numerical experiments. A field with non-convex contour (gray) is embedded within a real-world topographical map. The elevation profile (z-value) ranges from below 10m to more than 50m above sea-level. The field has a size of 28.6ha.}
\label{fig_data3d}
\end{figure}

\begin{figure}
\centering
\includegraphics[width=8.8cm]{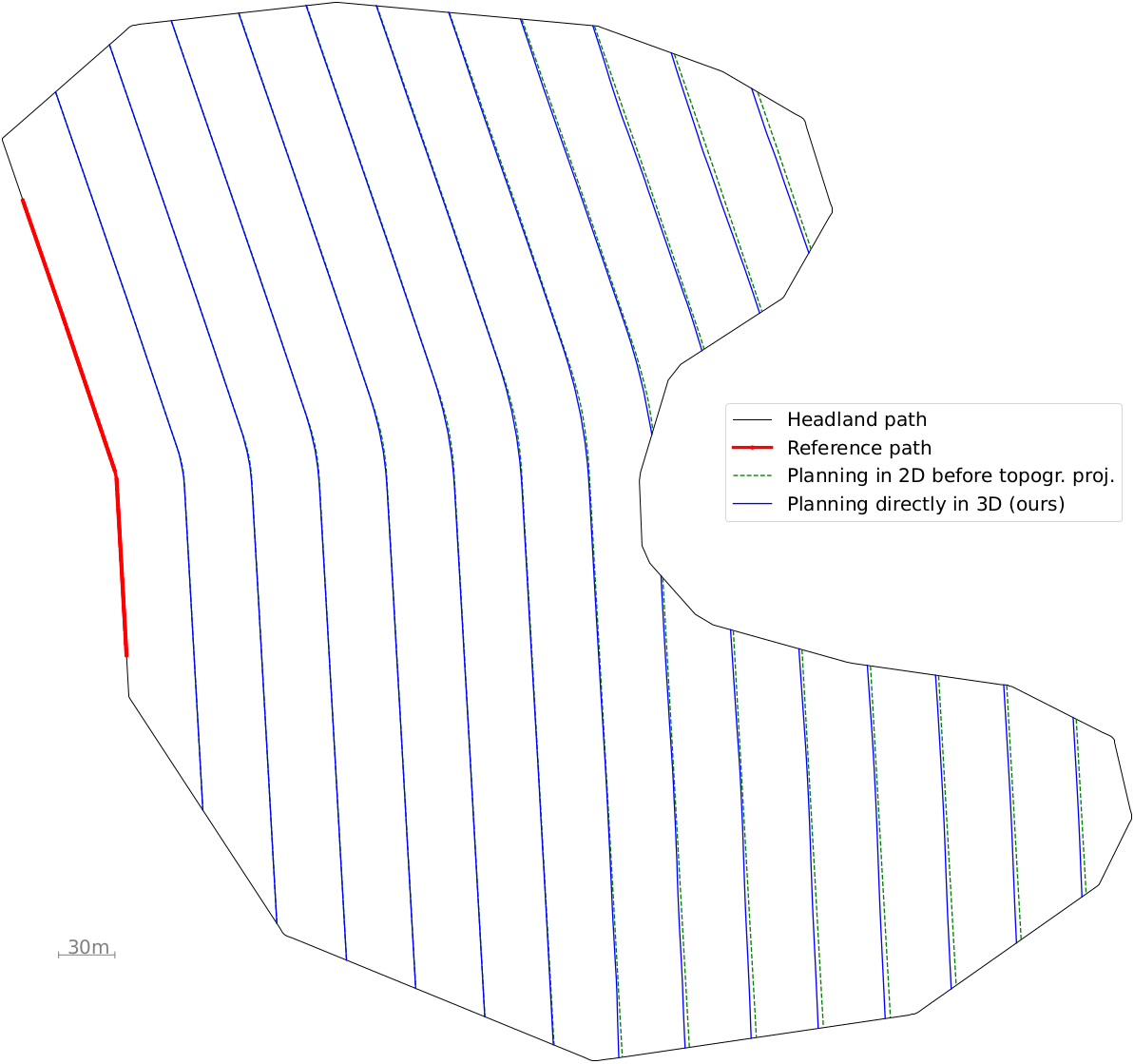}
\caption{Creation of mainfield lanes: bird's eye view comparison of the methods of (i) hierarchically planning first freeform lanes in 2D according to \cite{plessen2021freeform} before their projections onto the terrain, and (ii) directly planning mainfield lanes in 3D according to the Algorithm of Sect. \ref{subsec_algo}. The working width and height are $w=36$m and $h=2$m. The maximum lateral deviation between the results of the 2 methods is 2.83m (see also the zoom-in in Fig. \ref{fig_heps0d1zoom}).}
\label{fig_heps0d1}
\end{figure}

\begin{figure}
\centering
\includegraphics[width=8.8cm]{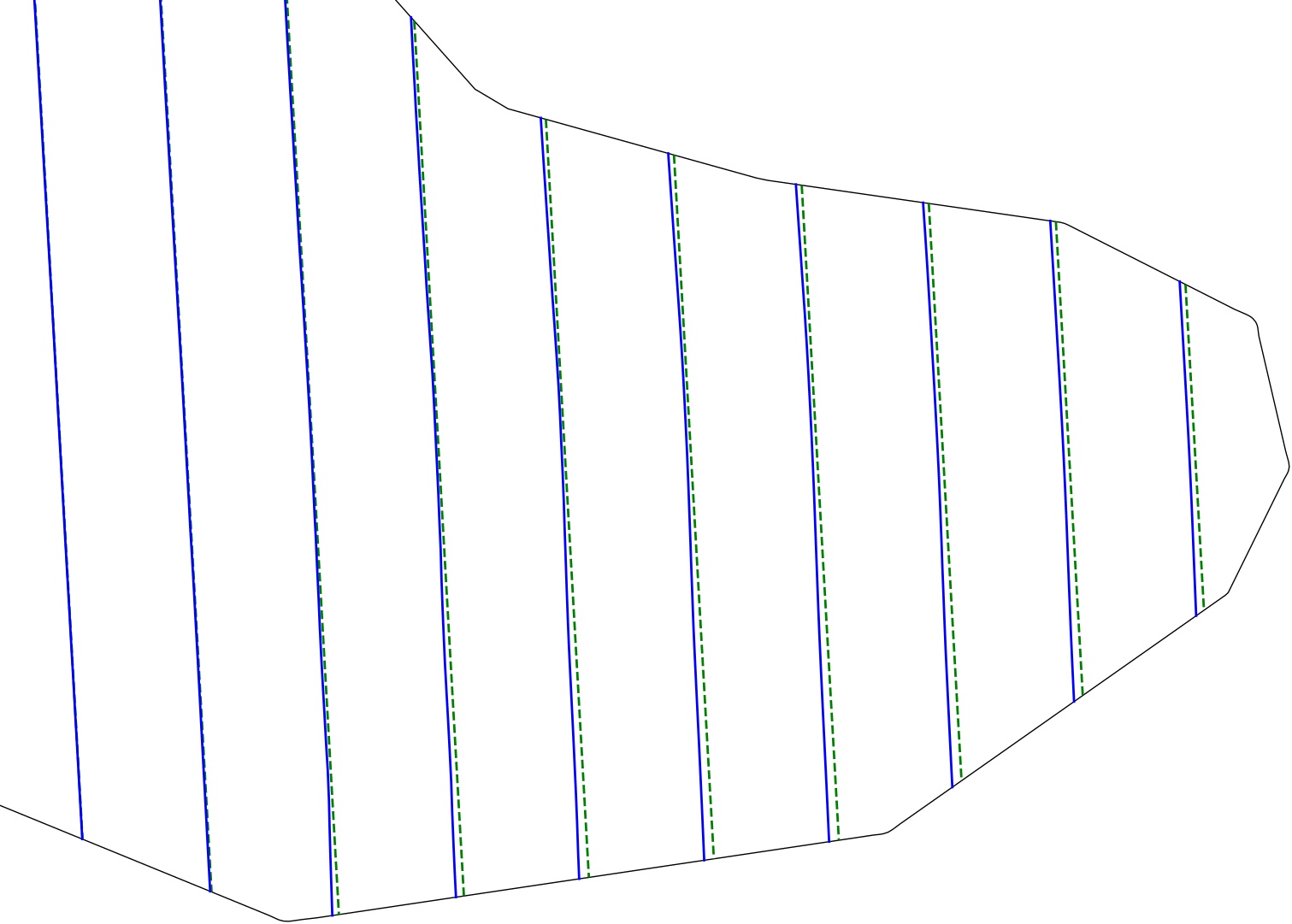}
\caption{Zoom-in into the south-eastern part of Fig. \ref{fig_heps0d1}. The maximum lateral deviation between the results of the 2 methods (dashed and solid) is 2.83m.}
\label{fig_heps0d1zoom}
\end{figure}

\begin{figure}
\centering
\includegraphics[width=8.8cm]{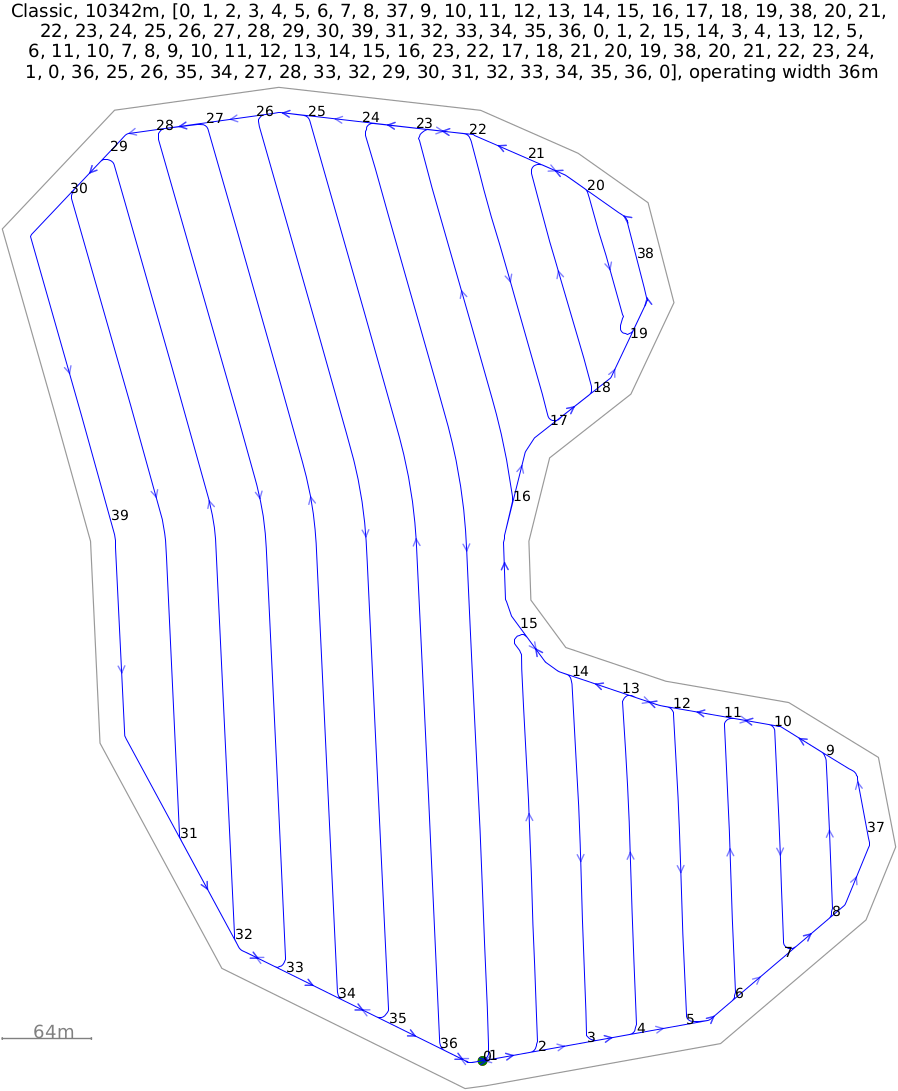}
\caption{Full area coverage path planning result. For the mainfield lanes generated by the 3D-planning approach in Fig. \ref{fig_heps0d1}, a sequence of mainfield lane traversals is assigned before transitions between mainfield lanes and headland path are smoothed according to \cite{plessen2025smoothing}. The field entry and exit location is indicated by the green 0-node.}
\label{fig_full}
\end{figure}

\begin{figure}
\centering
\includegraphics[width=3.0cm]{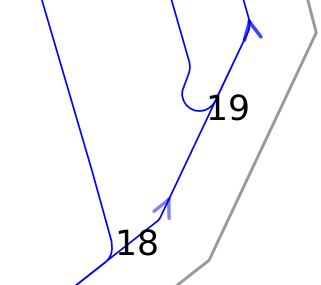}
\caption{A zoom-in into part of Fig. \ref{fig_full}.}
\label{fig_zoom}
\end{figure}

Data used for numerical experiments are illustrated in Fig. \ref{fig_data3d}. The results of 2 different methods for the generation of adjacent mainfield lanes while accounting for 3D terrain data are visualized in Fig. \ref{fig_heps0d1}. A zoom-in in Fig. \ref{fig_heps0d1zoom} further highlights the differences of the 2 methods. Only the solution resulting from planning directly in 3D avoids the spray overlap or spray gaps issues that are associated with 2D-planning before path projections onto the topographical map, as outlined in Sect. \ref{sec_probformul}. The full 3D terrain-following area coverage path planning result is shown in Fig. \ref{fig_full}. A zoom-in is in provided in Fig. \ref{fig_zoom} to highlight smooth transitions between mainfield lanes and the headland path.

\section{Discussion and Outlook\label{sec_discussion}}

Terrain-following area coverage planning in 3D is related to 2D freeform path fitting methods (e.g., \cite{plessen2021freeform}). However, the complexities stemming from the change to non-planar terrain are significant. They result in eqs. \eqref{eq_1stset_1st}-\eqref{eq_f_2}, \eqref{eq_2ndset_1st}-\eqref{eq_2ndset_hproj}, and iteration \eqref{eq_beta_sear} in combination with \eqref{eq_what}-\eqref{eq_2ndset_hproj} and \eqref{eq_omega}. The entire look-up table generation step from Sect. \ref{subsec_ztable} and hyperparameters in Table \ref{tab_hyperparam} are also not required for 2D.

The method from \cite{hameed2016side}, which was highlighted in the Introduction, could in general be adapted for the 3D problem discussed in this paper. This adaptation would encompass, (i) projecting the entire terrain locally orthogonal by a distance $h$, before then (ii) treating this projected surface as the new field surface and applying the cylindrical method from \cite{hameed2016side}. However, this projection step is in general not  straightforward, since dependent on interpolation spacing and working height easily jaggedness is introduced into the new surface, whose correction is not straightforward. Furthermore, the projection step is inefficient since in general (without the introduction of additional heuristics and hyperparameters) the \emph{entire} field surface would have to be projected, before the cylindrical method can be applied. 

There are 3 main avenues for future work. First, headland path edges and headland-to-mainfield lane transitions are currently smoothed before being vertically projected onto the terrain. For an improved method it is planned to use a \emph{3D dynamical tractor model} and adapt the method \cite{plessen2025smoothing} to calculate these transitions. Second, the modeling of terrain data can potentially be improved. Instead of using look-up tables, a polynomial fit  may be used (e.g., \cite{blane20023l,carr2001reconstruction}). A trade-off between potentially improved computational efficiency and loss of 3D terrain modeling accuracy is expected. Third, alternative heuristics for path tangential interpolation spacing (see Sect. \ref{subsec_dspacing}) may be reviewed to refine performance guarantees.


\section{Conclusion\label{sec_conclusion}}

An algorithm for the generation of 3D terrain-following area coverage path planning was presented. Multiple adjacent paths are generated that are (i) locally apart from each other by a distance equal to the working width of a machinery (e.g., the boombar width for agricultural spraying applications), while (ii) at the same time floating at each sampling point at a projection distance equal to a specific height  above the terrain (e.g. the boom height or nominal distance of nozzles above the ground). 

The complexities of the algorithm in comparison to its 2D equivalent are highlighted that arise due to the introduction of the third dimension. These include in particular, a look-up table generation using an Inverse Distance Weighting (IDW)-approach to generate elevation data over a uniform interpolation grid and a local search.

%
%
%
%
%

\nocite{*}
\bibliographystyle{ieeetr}
\bibliography{myref}

\end{document}